# MixiMotion: One-Step Text-to-Motion Generation via Asymmetric Set Distillation

Hung Dinh[1 [0009−0003−6278−1092] †], Binh Mai[1 [0009−0009−9957−8702] †], Tran Quoc Bao Le[1 [0009−0001−9024−5858]], Lam Nguyen[1 [0009−0005−3550−9864]], and Cong Tran[1 [0000−0001−9467−4978] ⋆]

Posts and Telecommunications Institute of Technology, Hanoi, Vietnam
{dinhhung15082004, k100iltqbao, hailam04.work}@gmail.com, {binhmd, congtt}@ptit.edu.vn
[†]These authors contributed equally.

**Abstract.** Iterative text-to-motion generation delivers high-quality and semantically aligned motions but requires multiple network evaluations, resulting in substantial inference latency. We present **MixiMotion**, a strict one-step text-to-motion generation framework based on offline set distillation. Instead of distilling a single teacher trajectory for each text prompt, MixiMotion constructs an offline bank of multiple teacher motions and aligns teacher and student sample sets through **asymmetric bidirectional matching**. The teacher-to-student direction promotes coverage of diverse teacher-supported motions, while the student-to-teacher direction suppresses unsupported generations. We further introduce differentiable decoded-space kinematic supervision to complement normalized representation matching with constraints in the decoded motion space. At inference, MixiMotion generates a complete motion sequence with a single network evaluation, without teacher queries, iterative sampling, or candidate ranking. On ViMoGen, MixiMotion achieves a semantic alignment score of 0.835, outperforming the evaluated one-step baselines and approaching the 0.858 score of its 50-step HY-Motion-1.0-Lite teacher. In blinded human evaluation, MixiMotion obtains an overall rating of 4.33, compared with 4.50 for the teacher, while outperforming the evaluated one-/few-step baselines. Meanwhile, generation latency is reduced from 829.58 ms to 9.30 ms, corresponding to an 89.2× speedup. These results demonstrate an effective quality–efficiency trade-off for strict one-step text-to-motion generation.
Project page: https://miximotion.ptitai.org/

## 1 Introduction

Text-to-motion generation aims to synthesize realistic 3D human motion from natural-language descriptions. Recent diffusion- and flow-based models have substantially improved motion quality and text–motion alignment [18,2,20], but typically require tens to hundreds of iterative network evaluations. This high sampling cost limits their applicability to latency-sensitive scenarios such as interactive animation, games, and online content creation.

⋆ Corresponding author: Cong Tran (congtt@ptit.edu.vn).

Compressing such iterative generators into a strict one-step model is non-trivial. First, text-conditioned motion is inherently multimodal: the same prompt can correspond to multiple semantically valid motions, making fixed teacher–student correspondence overly restrictive and prone to averaging across valid modes. Second, matching motions only in a normalized representation does not guarantee visually plausible decoded motion, since small representation errors may manifest as noticeable kinematic artifacts. These challenges suggest that effective one-step distillation should preserve multiple teacher-supported solutions while explicitly constraining motion plausibility after decoding.

We introduce **MixiMotion**, a strict one-step text-to-motion framework based on **offline asymmetric set distillation**. For each text prompt, a frozen multi-step teacher generates multiple motion candidates that are cached once in an offline teacher bank. The student then produces multiple one-step samples from independent noise and is trained by asymmetric bidirectional set matching: teacher-to-student matching encourages coverage of diverse teacher-supported motions, whereas student-to-teacher matching suppresses unsupported generations. We further complement representation-space matching with differentiable supervision in the decoded motion space. Importantly, the teacher is required only for offline bank construction and is never queried during student optimization or inference.

Experiments on ViMoGen demonstrate that MixiMotion substantially closes the quality gap between one-step and iterative generation. It achieves a semantic alignment score of 0.835, compared with 0.858 for its 50-step HY-Motion-1.0-Lite teacher, while outperforming the evaluated one-step baselines. In blinded human evaluation, MixiMotion obtains an overall rating of 4.33, compared with 4.50 for the teacher. At the same time, it reduces generation latency from 829.58 ms to 9.30 ms, yielding an 89.2× speedup.

Our contributions are summarized as follows:

- We formulate strict one-step text-to-motion distillation as **offline multi-sample set matching**, avoiding fixed correspondence between independently generated teacher and student motions.
- We propose an **asymmetric bidirectional matching objective** that separately controls teacher-mode coverage and suppression of unsupported student generations.
- We introduce **differentiable decoded-space kinematic supervision** to complement normalized representation matching and improve motion plausibility.
- We provide a comprehensive evaluation of semantic alignment, human-rated quality, ablations, and inference efficiency, showing that MixiMotion approaches the quality of a 50-step teacher with a single network evaluation.

## 2 Related Work

### 2.1 Text-to-motion generation.

Text-to-motion generation has evolved from autoregressive approaches, including T2M-GPT [22] and MotionGPT [9], to diffusion- and flow-based models such as MDM [18], MotionDiffuse [23], and MLD [2]. Subsequent methods improve generation through retrieval, sampling, or architectural advances, including ReMoDiffuse [24], StableMoFusion [8], Motion Mamba [25], and the large-scale flow-matching HY-Motion-1.0 [20]. Despite strong generation quality, these approaches generally rely on iterative sampling with multiple network evaluations. In contrast, MixiMotion targets strict one-step generation, producing a complete motion sequence with a single student network evaluation.

### 2.2 Distillation and accelerated generative modeling.

Knowledge distillation [7] has been widely used to transfer expensive generative models into more efficient students. For diffusion and flow models, progressive distillation [16], consistency models [17], and rectified flow [13] reduce sampling cost through trajectory compression or simplification, while Distribution Matching Distillation [21] and Score Identity Distillation [26] perform distribution-level alignment without fixed sample-wise correspondence. One-step distillation has also been explored in other domains, such as SwiftAudio [15]. For human motion, MotionLCM [3], MotionPCM [10], and MotionHiFlow [11] accelerate generation through consistency training or shortened flow-based sampling. Unlike these trajectory-oriented approaches, MixiMotion uses the teacher only offline to construct multiple cached motions per prompt and trains directly on these outputs, requiring neither teacher evaluation nor iterative trajectory simulation during student optimization or inference.

### 2.3 Set-based matching for multimodal generation.

Because conditional generation is inherently multimodal, fixed one-to-one supervision can be restrictive when multiple outputs are valid. IMLE [12] addresses this through nearest-sample matching, while Chamfer-style objectives [4] compare unordered sets without predefined correspondences; MoFlow [5] further applies one-step IMLE-style distillation to multimodal human trajectory forecasting. MixiMotion extends this sample-set perspective to language-conditioned full-body motion synthesis by matching multiple cached teacher motions and student samples with an asymmetric bidirectional objective: teacher-to-student matching promotes coverage of teacher-supported motions, whereas student-to-teacher matching suppresses unsupported generations. This set-level supervision is further complemented by differentiable decoded-space kinematic supervision to improve motion plausibility.

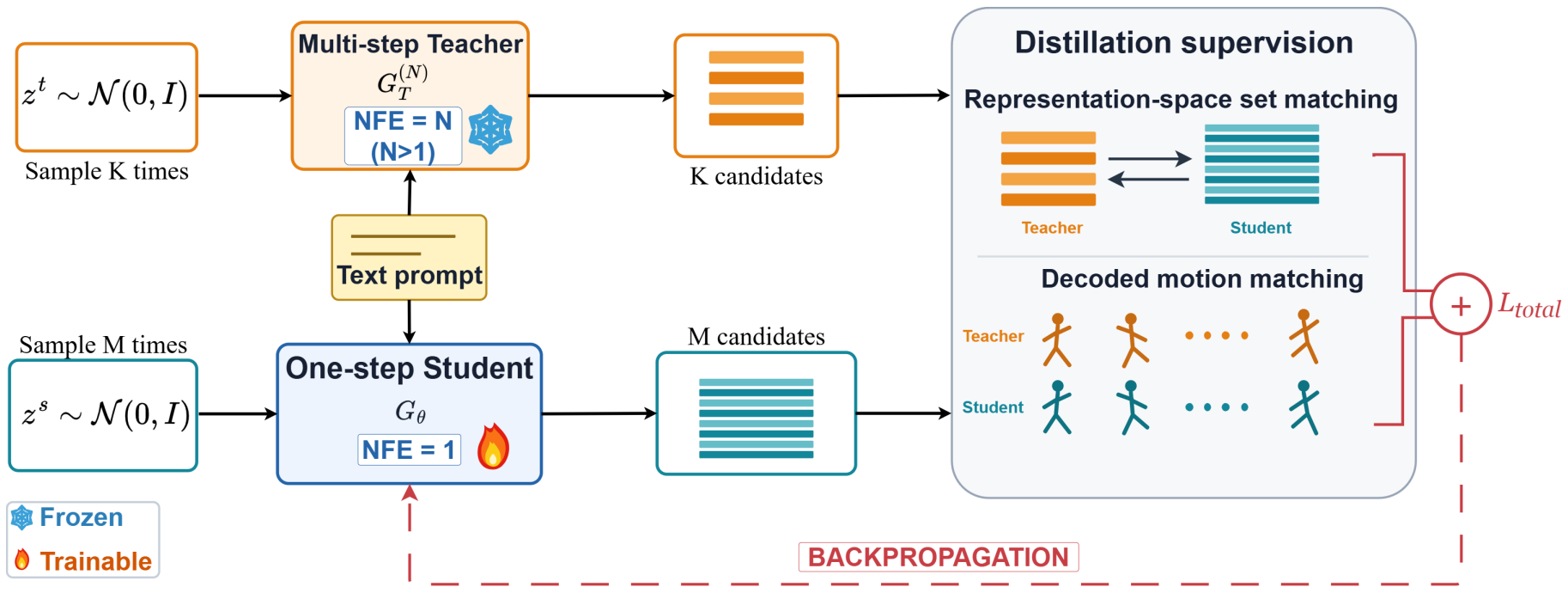

**Fig. 1.** Overview of MixiMotion. A frozen multi-step teacher generates $K$ motion candidates per prompt to construct an offline teacher bank, while the one-step student generates $M$ candidates from independent noise samples. The teacher and student sets are aligned using asymmetric bidirectional matching in the normalized representation and differentiably decoded motion spaces. Only the student is optimized during training.

# 3 Method

Figure 1 provides an overview of the proposed MixiMotion framework. A frozen multi-step teacher is used offline to construct a multi-candidate motion bank for each text prompt, while the one-step student generates a set of motion candidates from independently sampled noise. Training aligns the two sets using asymmetric bidirectional matching in both representation and decoded motion spaces, with gradients propagated only through the student.

## 3.1 Problem Formulation

We consider the problem of distilling an iterative text-to-motion generator into a strict one-step generator. Let $p$ denote a text prompt, $G_T^{(N)}$ a frozen teacher that generates a motion through an $N$-step iterative sampling process, and $G_\theta$ a one-step student parameterized by $\theta$. Thus, each teacher sample requires $N$ sequential network evaluations, whereas the student directly predicts the complete motion sequence with a single network evaluation.

Text-to-motion generation is inherently multimodal: a single prompt may admit multiple semantically correct and physically plausible motions. Therefore, enforcing a fixed correspondence between one student output and one teacher trajectory can provide overly restrictive supervision. Instead, we formulate distillation as matching two finite sets of motion samples conditioned on the same prompt.

### 3.2 Offline Multi-Candidate Teacher Bank

For each prompt $p$, the frozen teacher independently generates $K$ motion samples,

$$\mathcal{Y}_T(p) = \left\{ y_j = G_T^{(N)}(z_j^t, p) \right\}_{j=1}^{K}, \qquad z_j^t \overset{\text{i.i.d.}}{\sim} \mathcal{N}(0, I), \tag{1}$$

where $G_T^{(N)}$ denotes the complete teacher sampling process consisting of $N$ sequential network evaluations, and each $y_j \in \mathbb{R}^{T \times D}$ is represented in a normalized motion space with representation dimension $D$.

Constructing $K$ teacher candidates for a prompt therefore requires $K \times N$ teacher network evaluations. However, these samples are generated only once before student optimization and stored in an offline bank. Consequently, this iterative teacher cost is paid only during offline bank construction, and the teacher is never queried during student training or inference.

For the same prompt, the student draws $M$ independent noise samples and produces

$$\mathcal{X}_\theta(p) = \{ \hat{y}_i = G_\theta(z_i^s, p) \}_{i=1}^{M}, \qquad z_i^s \overset{\text{i.i.d.}}{\sim} \mathcal{N}(0, I). \tag{2}$$

The teacher and student noise variables are sampled independently. Consequently, no predefined one-to-one correspondence is assumed between $y_j$ and $\hat{y}_i$. Training instead compares the unordered sets $\mathcal{Y}_T(p)$ and $\mathcal{X}_\theta(p)$.

### 3.3 Robust Pairwise Motion Distance

To compare two motion representations, we use the robust penalty

$$\rho_\delta(e) = \begin{cases} \dfrac{e^2}{2\delta}, & |e| < \delta, \\ |e| - \dfrac{\delta}{2}, & |e| \geq \delta, \end{cases} \tag{3}$$

where $\delta > 0$ controls the transition between quadratic and linear behavior.

For a collection of valid frame indices $S \subseteq \{1, \ldots, \ell\}$, we define

$$d_\delta^S(a, b) = \frac{1}{|S|D} \sum_{t \in S} \sum_{d=1}^{D} \rho_\delta\left(a_{t,d} - b_{t,d}\right). \tag{4}$$

We use three complementary temporal resolutions. The complete valid sequence is

$$S_{\text{all}} = \{1, \ldots, \ell\}, \tag{5}$$

and the final-frame set is

$$S_{\text{end}} = \{\ell\}. \tag{6}$$

To describe coarse temporal evolution, we additionally choose $P$ uniformly distributed locations,

$$s_q = 1 + \text{round}\left(\frac{q(\ell - 1)}{P - 1}\right), \qquad q = 0, \ldots, P - 1, \tag{7}$$

where $2 \leq P \leq \ell$, and define

$$S_{\mathrm{traj}} = \{s_q\}_{q=0}^{P-1}. \tag{8}$$

For teacher sample $y_j$ and student sample $\hat{y}_i$, the corresponding pairwise costs are

$$C_{ji}^{\mathrm{lat}} = d_{\delta_{\mathrm{lat}}}^{S_{\mathrm{all}}}(y_j, \hat{y}_i), \tag{9}$$

$$C_{ji}^{\mathrm{end}} = d_{\delta_{\mathrm{lat}}}^{S_{\mathrm{end}}}(y_j, \hat{y}_i), \tag{10}$$

$$C_{ji}^{\mathrm{traj}} = d_{\delta_{\mathrm{lat}}}^{S_{\mathrm{traj}}}(y_j, \hat{y}_i). \tag{11}$$

The full-sequence term measures overall representation similarity, while the endpoint and sparse-trajectory terms give additional emphasis to the final state and coarse temporal evolution.

### 3.4 Asymmetric Bidirectional Set Matching

Because multiple motions can be valid for the same text prompt, imposing a fixed teacher–student correspondence is unnecessarily restrictive. Related nearest-neighbor and set-based matching strategies have also been used for distilling multimodal generative models, e.g., the IMLE-based distillation framework of MoFlow [5]. Motivated by this general principle, we match the teacher and student samples directly at the set level using an asymmetric bidirectional objective.

Let

$$\boldsymbol{\alpha} = (\alpha_{\mathrm{t2s}}, \alpha_{\mathrm{s2t}}), \qquad \alpha_{\mathrm{t2s}}, \alpha_{\mathrm{s2t}} \geq 0, \tag{12}$$

denote the directional matching weights. Here, $\alpha_{\mathrm{t2s}}$ controls the teacher-to-student matching term, which encourages coverage of the cached teacher samples, whereas $\alpha_{\mathrm{s2t}}$ controls the student-to-teacher matching term, which discourages student samples that are unsupported by the teacher bank.

For a pairwise cost matrix $C \in \mathbb{R}^{K \times M}$, we define the asymmetric bidirectional set-matching operator as

$$\mathcal{B}_{\boldsymbol{\alpha}}(C) = \alpha_{\mathrm{t2s}} \frac{1}{K} \sum_{j=1}^{K} \min_{1 \leq i \leq M} C_{ji} + \alpha_{\mathrm{s2t}} \frac{1}{M} \sum_{i=1}^{M} \min_{1 \leq j \leq K} C_{ji}. \tag{13}$$

The first term measures whether each teacher sample is represented by at least one nearby student sample and therefore promotes coverage of the finite teacher set. The second term measures whether each student sample is supported by at least one teacher sample and therefore penalizes unsupported generations.

The formulation is asymmetric because the two directions need not receive equal importance. In particular,

$$\alpha_{\mathrm{t2s}} \neq \alpha_{\mathrm{s2t}}$$

allows the trade-off between teacher-set coverage and suppression of unsupported student samples to be adjusted explicitly.

Equation (13) is a weighted asymmetric Chamfer-style reduction rather than a bijective assignment. Chamfer-based nearest-neighbor matching has also been employed in IMLE-based distillation for multimodal trajectory generation [5]. In contrast, our formulation explicitly separates teacher-to-student and student-to-teacher matching and assigns independent weights to the two directions, allowing teacher-set coverage and suppression of unsupported student samples to be controlled separately. Multiple teacher samples may select the same student sample, and multiple student samples may select the same teacher sample.

### 3.5 Decoded-Space Supervision

Representation-space similarity alone may not fully capture perceptually important motion properties after decoding. We therefore complement the representation-space losses with differentiable supervision defined on decoded motion features.

Let

$$F_{\text{dec}} : \mathbb{R}^{T \times D} \to \mathcal{Q} \tag{14}$$

be a fixed differentiable decoder from the normalized representation to a decoded motion space $\mathcal{Q}$.

For each feature type $r$ in a selected feature set $\mathcal{R}$, let

$$\Phi_r : \mathcal{Q} \to \mathcal{F}_r \tag{15}$$

be a differentiable feature map. We define

$$\psi_r(y) = \Phi_r(F_{\text{dec}}(y)) \,. \tag{16}$$

The feature set $\mathcal{R}$ may contain, for example, joint positions, root-motion quantities, selected body-part trajectories, or temporal differences of decoded features. Its precise instantiation depends on the motion representation and body decoder, while the set-distillation formulation itself is independent of a particular skeleton convention.

Let $\Omega_r(\ell)$ denote the valid scalar entries of feature $r$ for a sequence of length $\ell$. We define

$$C_{ji}^{(r)} = \frac{1}{|\Omega_r(\ell)|} \sum_{u \in \Omega_r(\ell)} \rho_{\delta_{\text{joint}}} \left( [\psi_r(y_j)]_u - [\psi_r(\hat{y}_i)]_u \right), \tag{17}$$

where $\delta_{\text{joint}} > 0$ is the robust-loss threshold associated with feature $r$.

The decoded-space pairwise cost is

$$C_{ji}^{\text{body}} = \sum_{r \in \mathcal{R}} \lambda_r C_{ji}^{(r)}, \qquad \lambda_r \geq 0. \tag{18}$$

This formulation is deliberately general: any differentiable decoded feature may be included without changing the set-matching objective.

### 3.6 Overall Set-Distillation Objective

The complete objective combines representation-space and decoded-space matching:

$$\begin{aligned}\mathcal{L}_{\text{total}} = \lambda_{\text{lat}}\, \mathcal{B}_{\boldsymbol{\alpha}}\left(C^{\text{lat}}\right) + \lambda_{\text{end}}\, \mathcal{B}_{\boldsymbol{\alpha}}\left(C^{\text{end}}\right) \\ + \lambda_{\text{traj}}\, \mathcal{B}_{\boldsymbol{\alpha}}\left(C^{\text{traj}}\right) + \lambda_{\text{body}}\, \mathcal{B}_{\boldsymbol{\alpha}}\left(C^{\text{body}}\right),\end{aligned} \tag{19}$$

where

$$\lambda_{\text{lat}}, \lambda_{\text{end}}, \lambda_{\text{traj}}, \lambda_{\text{body}} \geq 0. \tag{20}$$

The nearest-neighbor reduction is applied independently to each cost matrix. Therefore, different supervision components may select different nearest teacher–student pairs. This avoids imposing a single matching criterion across representation-space and decoded-space quantities.

### 3.7 Interpretation

The objective matches finite conditional sample sets rather than recovering the full teacher distribution. The teacher-to-student term promotes coverage of the $K$ cached teacher samples, whereas the student-to-teacher term discourages generations unsupported by this finite bank. Hence, a small $\mathcal{L}_{\text{total}}$ indicates agreement with the cached teacher support, not recovery of the complete conditional distribution. At inference, a single motion is generated as

$$\hat{y} = G_{\theta}(z, p), \qquad z \sim \mathcal{N}(0, I), \tag{21}$$

requiring one student evaluation with no teacher query or candidate ranking. Training and inference pseudocode is provided in the supplementary material.

## 4 Experiments

*Dataset.* We train our student model on ViMoGen [19]. For each training prompt, we construct an offline bank of teacher-generated motions, which is used throughout distillation.

*Compared Methods.* We compare MixiMotion with both one-step and multi-step text-to-motion generation methods. Among one-step methods, MotionLCM-1 [3] serves as a representative distilled baseline, while *Naive 1-step teacher* denotes HY-Motion-1.0-Lite [20] evaluated directly with a single denoising step, providing a controlled reference for the benefit of our set-distillation objective. For multi-step methods, we include our HY-Motion-1.0-Lite teacher together with representative diffusion-based baselines, namely MLD [2], MDM [18], and MotionDiffuse [23]. We additionally include MotionHiFlow [11] as a few-step baseline in experiments where corresponding results are available. Unless otherwise stated, we use the official implementations and their recommended inference settings for all competing methods.

We evaluate the methods from three complementary perspectives: text–motion semantic alignment, generation efficiency, and human perceptual preference.

### 4.1 Implementation Details

We use a frozen HY-Motion-1.0-Lite model [20] as the teacher. For each ViMoGen training prompt, we precompute an offline bank of $K = 4$ teacher motion samples; thus, the teacher is never queried during student optimization. Both the cached teacher motions and the student predictions remain in the native HY-Motion-1.0 motion representation, with sequence length $T = 120$ and feature dimension $D = 201$, avoiding any intermediate conversion to a separate motion representation.

The student reuses the HY-Motion-1.0-Lite MM-DiT architecture (460.0M generator parameters), is warm-started from the teacher, and maps Gaussian noise together with cached text features directly to a clean motion sequence in a single network evaluation. During training, we sample $M = 8$ student motions per prompt and optimize the asymmetric set objective in Eq. (19). We use asymmetric matching weights $\boldsymbol{\alpha} = (1, 0.20)$, assigning a stronger weight to teacher-to-student coverage than to student-to-teacher suppression.

Training uses AdamW [14] for 18 epochs with a constant learning rate of $8 \times 10^{-5}$. We maintain an exponential moving average of the student parameters and use the EMA model for all reported inference results. Full optimization hyperparameters, motion feature definitions, loss coefficients, decoded joint groups, and decoded-feature weights are provided in the supplementary material.

### 4.2 Semantic Alignment

*Protocol.* Following the SSAE protocol adopted by HY-Motion 1.0 [20], we randomly sample 2,000 ViMoGen prompts spanning six motion categories. Each prompt is decomposed into multiple yes/no semantic questions, and the rendered motion is evaluated by Qwen3-VL-30B-A3B-Instruct-FP8 [1] over all 120 frames, with temperature 0 and a fixed binary-response schema.

We report question-level correctness for each category. The Overall score is computed by pooling all question-level judgments across the 2,000 prompts, rather than averaging the six category scores; therefore, the Overall value need not equal their arithmetic mean.

*Results.* MixiMotion achieves the best overall score among one-step methods, reaching 0.835, compared with 0.825 for MotionLCM-1 and 0.711 for the naive one-step HY-Motion baseline. It outperforms MotionLCM-1 in four of the six categories, while remaining competitive on Locomotion and Social Interactions. Despite using only one function evaluation, MixiMotion also matches or approaches the multi-step baselines MDM (0.833) and MLD (0.840), and remains only 2.3 points below the 50-step HY-Motion-1.0-Lite teacher (0.858).

### 4.3 Inference Efficiency

MixiMotion reduces the 50-step HY-Motion-1.0-Lite teacher to a single network evaluation, achieving an 89.2× generation speedup (829.58 ms $\rightarrow$ 9.30 ms) and a 6.75× end-to-end speedup. Detailed efficiency benchmarks and measurement protocols are provided in the supplementary material.

**Table 1.** Question-level semantic correctness on ViMoGen. Bold indicates the best result within each method category.

| Method | NFE | Daily↑ | Fit.↑ | Game↑ | Loco.↑ | Social↑ | Sports↑ | Overall↑ |
|---|---|---|---|---|---|---|---|---|
| *One-step methods* | | | | | | | | |
| MixiMotion (ours) | 1 | **0.842** | **0.721** | **0.886** | 0.944 | 0.903 | **0.806** | **0.835** |
| MotionLCM-1 [3] | 1 | 0.801 | 0.709 | 0.881 | **0.951** | **0.911** | 0.772 | 0.825 |
| Naive 1-step teacher [20] | 1 | 0.681 | 0.604 | 0.743 | 0.823 | 0.768 | 0.642 | 0.711 |
| *Multi-step methods* | | | | | | | | |
| Our teacher [20] | 50 | **0.874** | **0.738** | **0.901** | **0.958** | 0.918 | **0.823** | **0.858** |
| MLD [2] | 50 | 0.811 | 0.708 | 0.896 | 0.958 | **0.931** | 0.780 | 0.840 |
| MDM [18] | 50 | 0.820 | 0.724 | 0.865 | 0.950 | 0.927 | 0.752 | 0.833 |
| MotionDiffuse [23] | 1000 | 0.782 | 0.689 | 0.855 | 0.926 | 0.899 | 0.742 | 0.809 |

### 4.4 Human Evaluation

*Protocol.* We conduct a blinded human evaluation with 25 raters on 300 ViMoGen prompts per system. The generated motions from all compared methods are presented in randomized order without model identities, and raters assess each clip using five-point Likert scales for semantic alignment, naturalness, and overall quality. Table 2 reports mean±SE together with Krippendorff's $\alpha = 0.68$ [6]. Aggregate SEs are descriptive and are not used for paired significance tests. Further details on participant selection, rating criteria, and the evaluation interface are provided in the supplementary material.

**Table 2.** Blinded human ratings on 300 ViMoGen prompts per system (mean±SE). Bold and underline indicate the best and second-best results, respectively, for each evaluation criterion. Krippendorff's $\alpha = 0.68$.

| Method | Alignment↑ | Naturalness↑ | Overall↑ |
|---|---|---|---|
| MixiMotion (ours) | <u>$4.32 \pm 0.05$</u> | <u>$4.35 \pm 0.04$</u> | <u>$4.33 \pm 0.05$</u> |
| HY-Motion-1.0-Lite | $\mathbf{4.48 \pm 0.04}$ | $\mathbf{4.52 \pm 0.03}$ | $\mathbf{4.50 \pm 0.04}$ |
| MotionHiFlow | $4.18 \pm 0.06$ | $4.22 \pm 0.05$ | $4.20 \pm 0.06$ |
| MotionLCM-1 | $3.95 \pm 0.07$ | $4.01 \pm 0.06$ | $3.98 \pm 0.07$ |

*Results.* HY-Motion-1.0-Lite achieves the highest mean human ratings across all three criteria. MixiMotion obtains the second-highest scores, with an overall rating of 4.33 compared with 4.50 for the multi-step teacher, while outperforming the one-/few-step baselines in alignment, naturalness, and overall quality.

**Table 3.** Ablation of the proposed loss components and directional matching weights. The last two rows use the full loss configuration.

| Configuration | Semantic↑ |
|---|---|
| Full-sequence only | 0.820 |
| + endpoint | 0.827 |
| + trajectory | 0.830 |
| + endpoint + trajectory | 0.832 |
| + decoded supervision | 0.835 |
| Symmetric matching $(1, 1)$ | 0.829 |
| Asymmetric matching $(1, 0.20)$ | **0.835** |

### 4.5 Qualitative Results

Figure 2 presents qualitative comparisons between HY-Motion-1.0-Lite and MixiMotion under the same renderer, camera, and lighting conditions. The displayed prompts are randomly sampled from the evaluation set rather than manually selected based on output quality. These examples are intended to illustrate representative visual characteristics of the generated motions and should not be interpreted as a quantitative estimate of failure rates.

### 4.6 Ablation Study

We progressively add the proposed supervision terms while keeping all other training settings unchanged. We further compare symmetric and asymmetric bidirectional matching under the full loss configuration.

The endpoint, trajectory, and decoded-space terms progressively improve semantic alignment, with the full objective reaching 0.835. Under the same full loss configuration, asymmetric matching $(1, 0.20)$ also improves over symmetric weighting $(1, 1)$, supporting the use of a stronger teacher-to-student coverage term while retaining weaker student-to-teacher regularization.

## 5 Conclusion

We presented **MixiMotion**, a one-step text-to-motion generation framework based on offline asymmetric set distillation with decoded-space supervision. MixiMotion achieves an overall semantic score of 0.835, compared with 0.858 for the 50-step HY-Motion-1.0-Lite teacher, while outperforming the evaluated one-step baselines. Human evaluation further shows competitive perceptual quality, with an overall rating of 4.33 versus 4.50 for the teacher. At the same time, MixiMotion reduces generation latency from 829.58 ms to 9.30 ms, corresponding to an 89.2× speedup. These results demonstrate that set-based distillation provides an effective quality–efficiency trade-off for strict one-step text-to-motion generation.

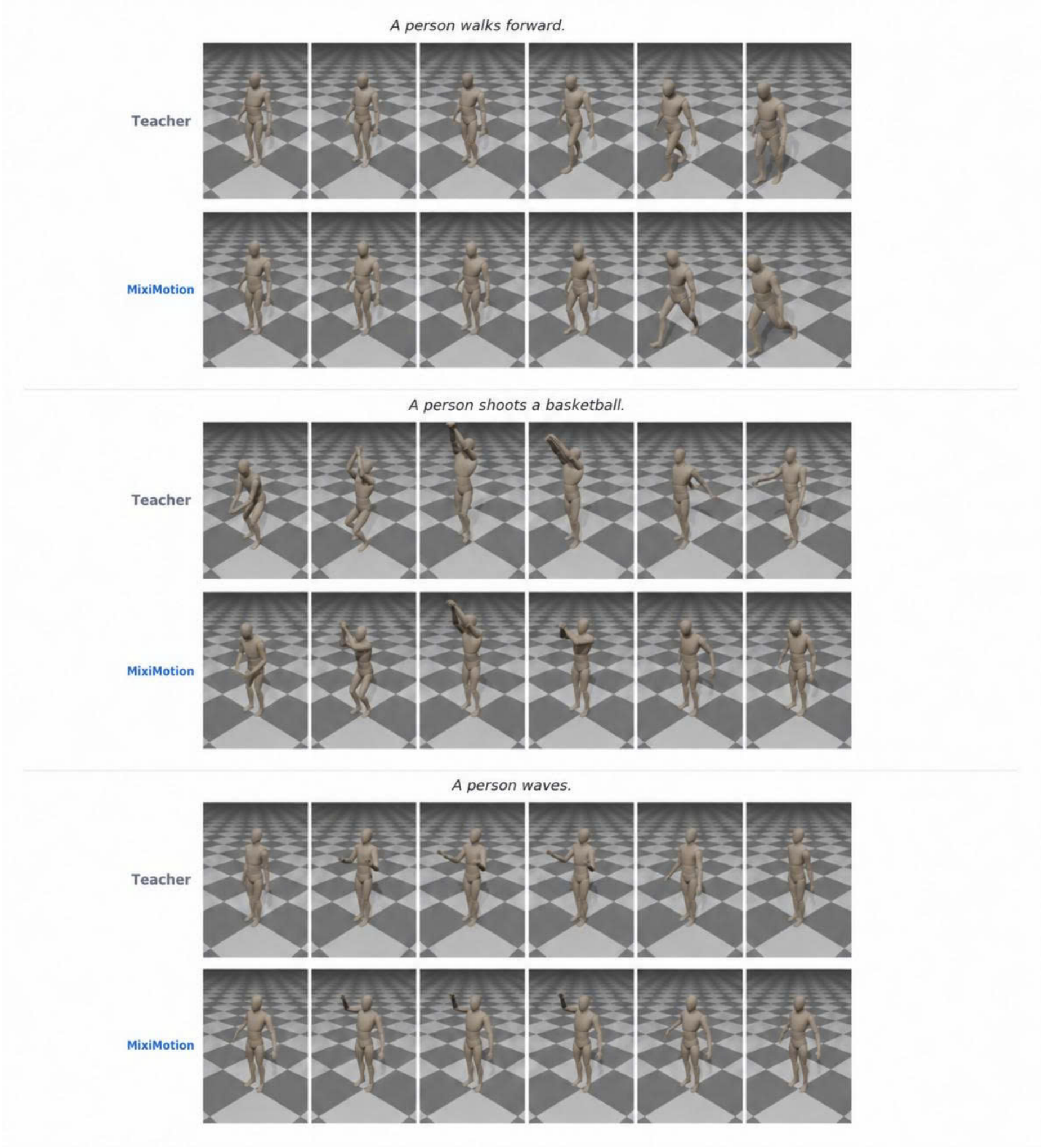

**Fig. 2.** Qualitative comparison on three randomly sampled evaluation instructions. HY-Motion-1.0-Lite (upper row) and MixiMotion (lower row) use the same renderer, camera, and lighting. The examples are sampled without manual selection based on generation quality.

# Supplementary Material: MixiMotion

Hung Dinh[1[0009−0003−6278−1092]†], Binh Mai[1[0009−0009−9957−8702]†], Tran Quoc Bao Le[1[0009−0001−9024−5858]], Lam Nguyen[1[0009−0005−3550−9864]], and Cong Tran[1[0000−0001−9467−4978]]

Posts and Telecommunications Institute of Technology, Hanoi, Vietnam
{dinhhung15082004, binhmai2205, k100iltqbao, hailam04.work}@gmail.com, congtt@ptit.edu.vn
[†]These authors contributed equally.

## 1 Implementation and Training Configuration

This section records the complete configuration used for the reported MixiMotion model. Unless otherwise stated, all experiments in the main paper use the settings below.

### 1.1 Teacher, Student, and Motion Representation

The teacher is a frozen HY-Motion-1.0-Lite model [2]. After exact-text filtering of the initial ViMoGen training prompt set, we generate an offline bank of $K = 4$ teacher motion samples for each retained prompt before student optimization. The resulting motions and frozen text features are cached and reused throughout training, so the teacher is never queried during student updates.

We retain the native HY-Motion-1.0 motion representation throughout distillation. Each motion frame is represented by a $D = 201$-dimensional vector containing root translation, 6D root orientation, 6D rotations for the 21 body joints, and an additional 66-dimensional posed-joint position feature. Both teacher and student motions therefore remain in the HY-Motion-1.0 motion space, without conversion to an alternative representation.

The student reuses the HY-Motion-1.0-Lite MM-DiT interface, with 460.0M generator parameters, and is initialized from the frozen teacher weights. Given Gaussian noise and cached text features, the student predicts a complete clean motion sequence in one network evaluation.

The reported configuration is

$$T = 120, \qquad D = 201, \qquad K = 4, \qquad P = 4, \qquad M = 8, \tag{S1}$$

where $T$ is the training sequence length, $K$ is the number of cached teacher motions per prompt, $P$ is the number of sparse trajectory knots, and $M$ is the number of independently sampled student motions per prompt during training. At inference, only one student sample is generated, i.e., $M = 1$.

### 1.2 Optimization

We optimize the student using AdamW [1] with

$$(\beta_1, \beta_2) = (0.9, 0.95), \tag{S2}$$

a constant learning rate of

$$8 \times 10^{-5}, \tag{S3}$$

weight decay 0.01, and gradient clipping with maximum norm 1.0. Training is performed for 18 epochs. Each optimization step uses four text prompts on a single GPU, and all training is carried out using bfloat16 precision.

During optimization, we maintain an exponential moving average (EMA) of the student parameters with decay 0.999. All evaluation and inference results reported in the main paper use this EMA copy.

### 1.3 Loss Weights and Robust-Penalty Thresholds

The asymmetric bidirectional set-matching weights are

$$\boldsymbol{\alpha} = (\alpha_{\mathrm{t2s}}, \alpha_{\mathrm{s2t}}) = (1, 0.20). \tag{S4}$$

The larger teacher-to-student weight prioritizes coverage of the teacher motion set, while the weaker student-to-teacher term penalizes student samples that are unsupported by the teacher bank.

The four matching components use

$$\lambda_{\mathrm{lat}} = 1, \qquad \lambda_{\mathrm{end}} = 0.10, \qquad \lambda_{\mathrm{traj}} = 0.12, \qquad \lambda_{\mathrm{dec}} = 1. \tag{S5}$$

Here $\lambda_{\mathrm{dec}}$ is the coefficient on decoded-space matching. Following the main-paper notation, $\lambda_{\mathrm{dec}} \equiv \lambda_{\mathrm{body}}$.

The latent-sequence, endpoint, and sparse-trajectory costs share

$$\delta_{\mathrm{lat}} = 0.35, \tag{S6}$$

while all decoded-feature robust penalties use

$$\delta_{\mathrm{joint}} = 0.04. \tag{S7}$$

### 1.4 Decoded-Feature Weights

Decoded-space supervision uses three joint groups:

$$\mathcal{G} = \{\mathrm{body}, \mathrm{foot}, \mathrm{key}\}. \tag{S8}$$

The *body* group contains all decoded joints, the *foot* group contains ankle and toe joints, and the *key* group contains the root, feet, head, and wrist joints.

Table S1 lists all nonzero decoded-feature weights. Any $(g, k)$ pair not listed has weight zero.

**Table S1.** Decoded-feature weights used for MixiMotion training. Unlisted $(g, k)$ pairs have weight zero.

| Group | Term | Symbol | Weight |
|---|---|---|---|
| Body | position | $\lambda_{\text{body},0}$ | 0.20 |
| | velocity | $\lambda_{\text{body},1}$ | 0.12 |
| | acceleration | $\lambda_{\text{body},2}$ | 0.04 |
| Foot | position | $\lambda_{\text{foot},0}$ | 0.30 |
| | velocity | $\lambda_{\text{foot},1}$ | 0.25 |
| | contact velocity | $\lambda_{\text{c}}$ | 0.32 |
| Key joints | position | $\lambda_{\text{key},0}$ | 0.18 |
| | velocity | $\lambda_{\text{key},1}$ | 0.12 |

## 2 Decoded Feature Maps and Loss Instantiation

This section specifies the decoded feature maps used to construct the decoded-space pairwise cost. Let

$$x = F_{\text{dec}}(y) \in \mathbb{R}^{\ell \times J \times 3} \tag{S9}$$

denote decoded 3D joint positions for a motion sequence $y$.

Temporal differences are unnormalized:

$$\Delta x_t = x_{t+1} - x_t, \qquad \Delta^2 x_t = \Delta x_t - \Delta x_{t-1}, \tag{S10}$$

with no division by $\Delta t$.

For joint group $g$, the decoded feature maps are

$$\Phi_{\text{pos},g}(x)_t = x_{t,g}, \tag{S11}$$

$$\Phi_{\text{vel},g}(x)_t = \Delta x_{t,g}, \tag{S12}$$

$$\Phi_{\text{acc,body}}(x)_t = \Delta^2 x_{t,\text{body}}. \tag{S13}$$

The contact-velocity term is teacher-masked. Let $h_{t,f}$ denote the height of foot $f$ above the lowest foot height in the clip. We define

$$c_{t,f}(y) = \mathbf{1}[h_{t,f}(y) < h_{\text{c}}], \qquad h_{\text{c}} = 0.05. \tag{S14}$$

For a student decoded motion $\hat{x}$, the contact feature is

$$\Phi_{\text{c}}(\hat{x}, y)_{t,f} = c_{t,f}(y)\, \|\Delta \hat{x}_{t,f}\|_2. \tag{S15}$$

Only the student side of $\Phi_{\text{c}}$ receives gradients.

Let $\Omega_r(\ell)$ denote the valid scalar entries of decoded feature $r$ for a sequence of length $\ell$, as in the main paper. Let $\mathcal{R}$ denote the set of active decoded features listed in Table S1. For teacher motion $y_j$ and student motion $\hat{y}_i$, the decoded pairwise cost is

$$C_{ji}^{\text{body}} = \sum_{r \in \mathcal{R}} \lambda_r C_{ji}^{(r)}, \tag{S16}$$

where each $C_{ji}^{(r)}$ uses the decoded robust penalty with $\delta_{\text{joint}}$ given in Section 1.3.

# 3 Offline Teacher Bank, Data Filtering, and Computational Cost

## 3.1 Offline Teacher Bank

Let $\mathcal{P}_{\text{raw}}$ denote the initial ViMoGen training text set, which contains

$$|\mathcal{P}_{\text{raw}}| = 50{,}000 \tag{S17}$$

prompts. Before teacher-bank construction, we apply the exact-text filter described in Section 3.2 and obtain the filtered training prompt set

$$\mathcal{P} = \text{Filter}(\mathcal{P}_{\text{raw}}). \tag{S18}$$

Only prompts in $\mathcal{P}$ are passed to the frozen HY-Motion-1.0-Lite teacher. The teacher uses $N = 50$ sampling NFEs, and we generate $K = 4$ teacher samples for every retained prompt. Therefore, constructing the bank requires

$$KN = 200 \tag{S19}$$

teacher NFEs per retained prompt and

$$|\mathcal{P}|KN \tag{S20}$$

teacher NFEs in total. Since $|\mathcal{P}| \leq |\mathcal{P}_{\text{raw}}| = 50{,}000$, the total offline teacher cost satisfies

$$|\mathcal{P}|KN \leq 50{,}000 \times 4 \times 50 = 10^7. \tag{S21}$$

This teacher cost is paid once before student optimization; the bank is not rebuilt during training.

The cached motion bank stores

$$|\mathcal{P}|KTD \tag{S22}$$

motion scalars, excluding frozen text features.

## 3.2 Exact-Text Filter

Before teacher-bank construction, every prompt in $\mathcal{P}_{\text{raw}}$ whose normalized text exactly matches a ViMoGen evaluation prompt is removed. Normalization consists of whitespace collapse and case folding. The retained prompts form $\mathcal{P}$, which is then used both for offline teacher-bank construction and subsequent student optimization. This filter certifies zero exact-string overlap with the evaluation set; it does not remove paraphrases or near-duplicates.

### 3.3 Pairwise Cost Construction and Complexity

For each training prompt, the student generates $M = 8$ one-step motion samples. Pairwise comparison against the $K = 4$ cached teacher samples produces four $K \times M$ cost matrices:

$$\begin{aligned} C_{ji}^{\mathrm{lat}} &= d_{\delta_{\mathrm{lat}}}^{S_{\mathrm{all}}}(y_j, \hat{y}_i), \\ C_{ji}^{\mathrm{end}} &= d_{\delta_{\mathrm{lat}}}^{S_{\mathrm{end}}}(y_j, \hat{y}_i), \\ C_{ji}^{\mathrm{traj}} &= d_{\delta_{\mathrm{lat}}}^{S_{\mathrm{traj}}}(y_j, \hat{y}_i), \\ C_{ji}^{\mathrm{body}} &= \sum_{r \in \mathcal{R}} \lambda_r C_{ji}^{(r)}. \end{aligned} \tag{S23}$$

With the reported configuration, each matrix contains $KM = 32$ teacher–student pairs.

Representation-space terms require

$$O(KMD\ell) \tag{S24}$$

work per prompt, while decoded-space terms require

$$O(KM|\Omega|) \tag{S25}$$

for the valid decoded entries, where $|\Omega|$ denotes the aggregate number of valid scalar decoded-feature entries. Each asymmetric set reduction $\mathcal{B}_{\boldsymbol{\alpha}}$ is $O(KM)$ and is applied independently to the four matrices; consequently, different terms may select different teacher–student pairs $(j, i)$.

The resulting per-prompt training objective is

$$\begin{aligned} \mathcal{L}(p) = \lambda_{\mathrm{lat}} \mathcal{B}_{\boldsymbol{\alpha}}(C^{\mathrm{lat}}) + \lambda_{\mathrm{end}} \mathcal{B}_{\boldsymbol{\alpha}}(C^{\mathrm{end}}) \\ + \lambda_{\mathrm{traj}} \mathcal{B}_{\boldsymbol{\alpha}}(C^{\mathrm{traj}}) + \lambda_{\mathrm{dec}} \mathcal{B}_{\boldsymbol{\alpha}}(C^{\mathrm{body}}). \end{aligned} \tag{S26}$$

All coefficients in Eq. (S26) are defined once in Section 1.3.

Table S2 summarizes NFE accounting. After the offline bank has been constructed, teacher NFE is zero during student training and inference.

**Table S2.** NFE accounting for the reported run. Teacher NFE is paid once during offline bank construction.

| Stage | Teacher NFE | Student NFE |
|---|---|---|
| Bank, per prompt | $KN = 200$ | 0 |
| Bank, full filtered $\mathcal{P}$ | $\lvert\mathcal{P}\rvert KN \leq 10^7$ | 0 |
| One training prompt | 0 | $M = 8$ |
| Inference | 0 | 1 |

**Algorithm S1** Offline asymmetric set distillation.

**Require:** Raw prompt set $\mathcal{P}_{\text{raw}}$; frozen teacher $G_T^{(N)}$ and decoder $F_{\text{dec}}$; student $G_\theta$; EMA decay $\mu$

$\mathcal{P} \leftarrow \text{ExactTextFilter}(\mathcal{P}_{\text{raw}})$<br>
$\mathcal{D} \leftarrow \varnothing$<br>
**for** $p \in \mathcal{P}$ **do**<br>
    **for** $j = 1$ to $K$ **do**<br>
        $y_j \leftarrow G_T^{(N)}(z_j^t, p), \quad z_j^t \sim \mathcal{N}(0, I)$<br>
    **end for**<br>
    cache $\mathcal{Y}_T(p)$ and frozen text features of $p$ in $\mathcal{D}$<br>
**end for**<br>
warm-start $\theta$ from the frozen teacher; $\theta_{\text{EMA}} \leftarrow \theta$<br>
**for** minibatch $B \subset \mathcal{D}$ **do**<br>
    **for** $p \in B$ **do**<br>
        **for** $i = 1$ to $M$ **do**<br>
            $\hat{y}_i \leftarrow G_\theta(z_i^s, p), \quad z_i^s \sim \mathcal{N}(0, I)$<br>
        **end for**<br>
        **for** $j = 1$ to $K$, $i = 1$ to $M$ **do**<br>
            construct $C_{ji}^{\text{lat}}$, $C_{ji}^{\text{end}}$, $C_{ji}^{\text{traj}}$, and $C_{ji}^{\text{body}}$ using Eq. (S23)<br>
        **end for**<br>
        compute $\mathcal{L}(p)$ using Eq. (S26)<br>
    **end for**<br>
    AdamW update on $\sum_{p \in B} \mathcal{L}(p)$<br>
    $\theta_{\text{EMA}} \leftarrow \mu\theta_{\text{EMA}} + (1 - \mu)\theta$<br>
**end for**<br>
**return** $G_{\theta_{\text{EMA}}}$

## 4 Training and Inference Procedures

Algorithm S1 summarizes exact-text filtering, offline teacher-bank construction, and student optimization. The raw ViMoGen training prompt set is filtered before any teacher sampling. Hyperparameters and loss weights are defined in Sections 1–3 and are not repeated in the algorithm except where needed to clarify the computation.

Inference uses no teacher query, no iterative solver, and no ranking over the $M$ samples used during training.

## 5 Subjective Evaluation Details

We describe the blinded Likert protocol and the assessment interface.

### 5.1 Participant Selection

We recruited 25 participants familiar with computer graphics or motion synthesis, with full proficiency in English reading, so that ViMoGen prompts could be interpreted. Participants watched each clip in full, in a quiet setting, and could replay a clip before scoring.

**Algorithm S2** One-step MixiMotion sampling.

**Require:** EMA student $G_{\theta_{\text{EMA}}}$; prompt $p$; length $\ell$; frozen text encoders and decoder $F_{\text{dec}}$

encode $p$ and build the validity mask of length $\ell$

sample $z \sim \mathcal{N}(0, I)$

$\hat{y} \leftarrow G_{\theta_{\text{EMA}}}(z, p)$ ▷ 1 student NFE

$x \leftarrow F_{\text{dec}}(\hat{y})$

**return** $x$

### 5.2 Evaluation Protocol

The study is *blinded*, not double-blind: model identities and left–right order are hidden from raters; the experimenters know the mapping. Clips from MixiMotion, HY-Motion-1.0-Lite, MotionHiFlow, and MotionLCM-1 are shown in randomized order. Each system is scored on 300 ViMoGen prompts. Raters use a 5-point Likert scale (1: Bad to 5: Excellent). For each unlabeled clip, evaluators score three criteria:

- **Semantic alignment:** How well the generated motion matches the text prompt.
- **Naturalness:** How human-like the motion is, including posture, contact, and absence of obvious artifacts such as skating or implausible poses.
- **Overall quality:** The clip as a whole, combining alignment and naturalness.

*Criteria for semantic alignment.*

- 5 – Excellent: the motion fully matches the prompt.
- 4 – Minor mismatch in timing or a secondary action.
- 3 – The main action is recognizable but incomplete or incorrect in part.
- 2 – Multiple prompt components missing or wrong.
- 1 – Completely unrelated to the prompt.

*Criteria for naturalness.*

- 5 – Excellent: human-like motion with stable contacts.
- 4 – Slightly stiff or a small contact artifact.
- 3 – Clearly synthetic but recognizable as the intended action.
- 2 – Strong skating, popping, or implausible poses.
- 1 – Broken or unrecognizable motion.

*Criteria for overall quality.*

- 5 – Excellent.
- 4 – Good, with only slight issues.
- 3 – Acceptable but clearly synthetic.
- 2 – Poor.
- 1 – Unusable.

We report mean±standard error over prompts and Krippendorff's $\alpha = 0.68$ in the main paper. Aggregate SEs are descriptive and are not used for paired significance tests. The study does not include participant-level inferential tests.

### 5.3 Assessment Interface

The evaluation was performed via a dedicated web-based interface, as shown in Figures S1 and S2. The interface provided the prompt text and allowed participants to replay each unlabeled clip before submitting Likert ratings. Model names were not displayed on the blinded page. Ratings could be exported as JSON.

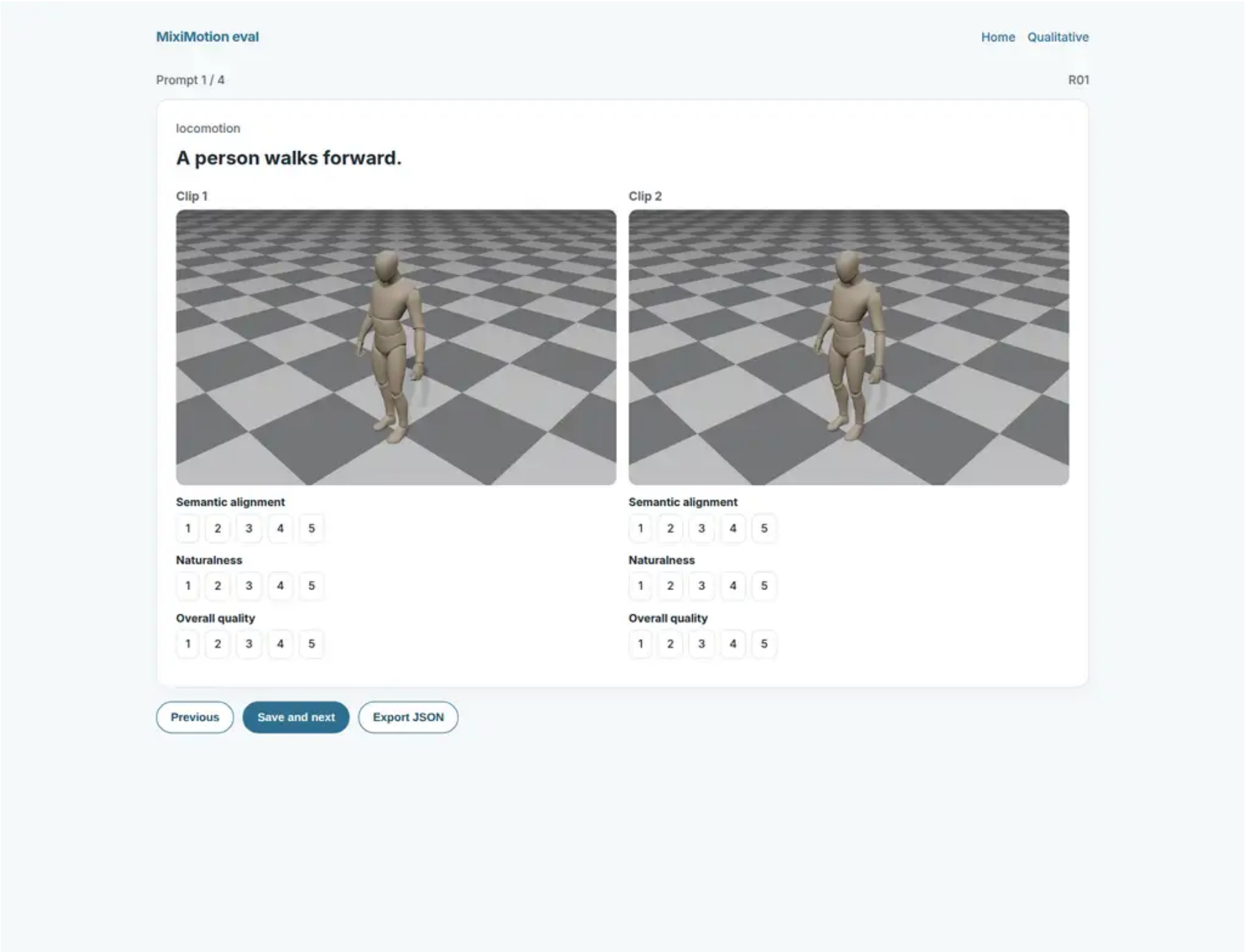

**Fig. S1.** Blinded rater interface. Each prompt shows unlabeled clips. Raters score semantic alignment, naturalness, and overall quality on five-point Likert scales.

## 6 Detailed Inference Efficiency Evaluation

### 6.1 Evaluation Protocol

We evaluate inference efficiency on a single NVIDIA RTX 5090 using BF16 precision, batch size 1, and 120-frame motion sequences. For each method, we perform 20 warm-up iterations followed by 100 synchronized runs. We report the number of network function evaluations (NFE), model-native motion-generation latency, and end-to-end (E2E) latency. The latter additionally includes text conditioning and motion decoding.

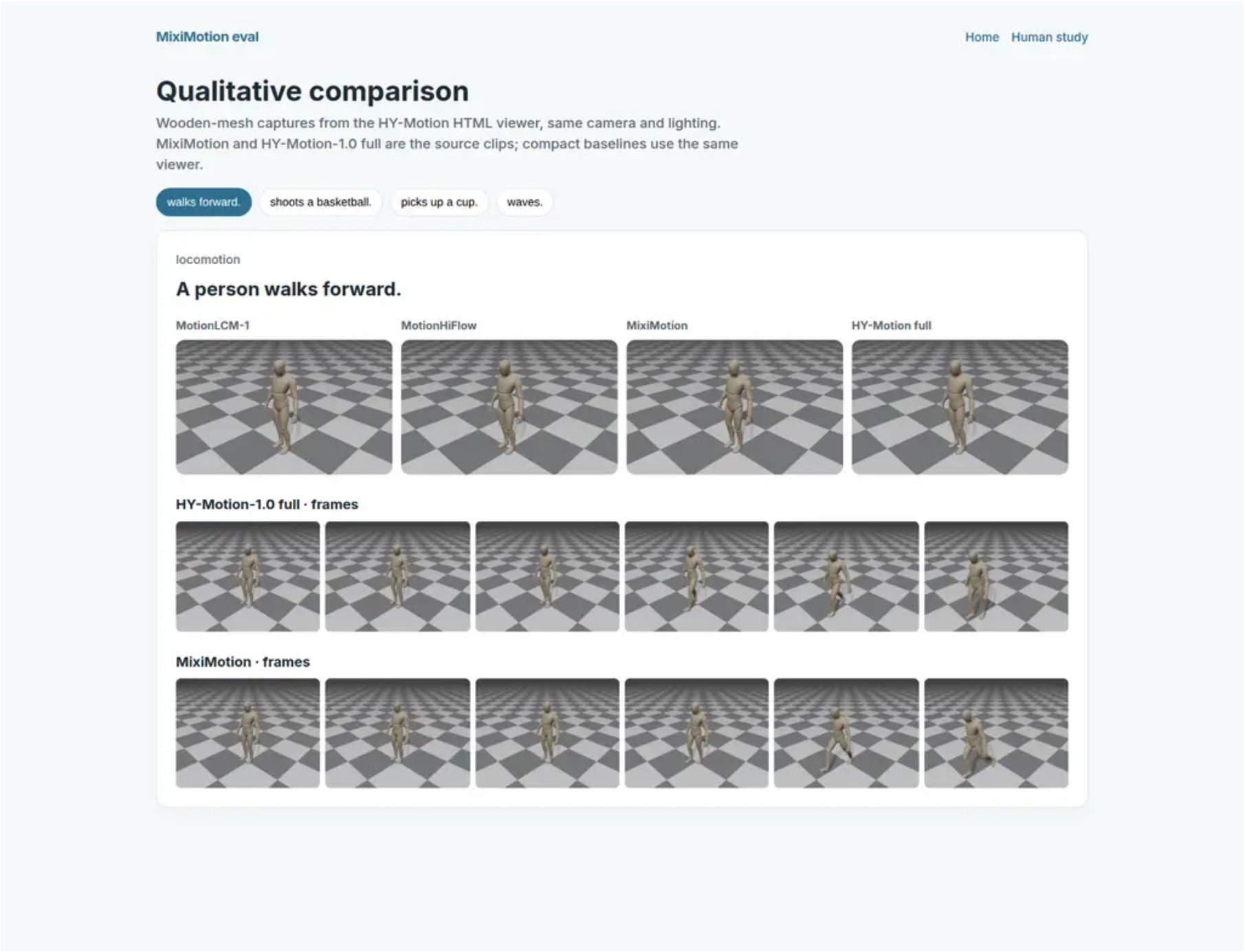

**Fig. S2.** Qualitative gallery. Labeled clips for four systems and six-frame strips for HY-Motion-1.0-Lite and MixiMotion.

We compare MixiMotion with the multi-step HY-Motion-1.0-Lite teacher, the few-step MotionHiFlow baseline, and the one-step MotionLCM-1 baseline. The comparison with HY-Motion-1.0-Lite is particularly controlled because MixiMotion retains the same 460M-parameter generator architecture while replacing its 50-step iterative generation process with a single student evaluation.

## 6.2 Results and Discussion

MixiMotion reduces model-native generation latency from 829.58,ms for the 50-step HY-Motion-1.0-Lite teacher to 9.30,ms, yielding an 89.2× speedup. End-to-end latency decreases from 968.25,ms to 143.55,ms, corresponding to a 6.75× speedup.

The smaller E2E speedup relative to the generation-only speedup is expected because text conditioning and motion decoding constitute fixed overheads that are not reduced proportionally by replacing iterative sampling with one-step generation. Nevertheless, the controlled same-family comparison demonstrates that the substantial reduction in generation latency is obtained without shrinking the 460M-parameter generator.

MotionLCM-1 remains faster in absolute latency, but it uses a substantially smaller 26.5M-parameter backbone. Therefore, this comparison reflects differ-ences in both model capacity and sampling strategy. MixiMotion instead targets acceleration of a stronger HY-Motion-family generator while preserving its model capacity, reducing its sampling procedure from 50 network evaluations to one.

**Table S3.** Inference efficiency on a single RTX 5090 using BF16, batch size 1, and 120-frame outputs. Latencies are averaged over 100 synchronized runs after 20 warm-up iterations.

| Method | Params (M) | NFE | Gen. (ms)↓ | E2E (ms)↓ |
|---|---|---|---|---|
| *Controlled same-family comparison* | | | | |
| MixiMotion (ours) | 460.0 | 1 | **9.30±0.39** | **143.55±1.29** |
| HY-Motion-1.0-Lite | 460.0 | 50 | 829.58±3.19 | 968.25±3.69 |
| *Cross-model efficiency reference* | | | | |
| MotionLCM-1 | 26.5 | 1 | **3.04±0.24** | **13.94±0.47** |
| MotionHiFlow | 33.1 | 14 | 77.32±1.33 | 83.80±1.17 |